\documentclass[conference]{IEEEtran}
\IEEEoverridecommandlockouts
\usepackage{cite}
\usepackage{amsmath,amssymb,amsfonts}
\usepackage{algorithmic}
\usepackage{graphicx}
\usepackage{textcomp}
\usepackage{xcolor}
\def\BibTeX{{\rm B\kern-.05em{\sc i\kern-.025em b}\kern-.08em
    T\kern-.1667em\lower.7ex\hbox{E}\kern-.125emX}}

\begin{document}

\title{From Causal Pairs to Causal Graphs
}

% \author{\\
% \\
% \\
% \\
% \\
% \\
% }

\author{
\IEEEauthorblockN{Rezaur Rashid}
\IEEEauthorblockA{\textit{Department of Computer Science} \\
\textit{UNC Charlotte}\\
Charlotte, USA \\
mrashid1@uncc.edu}
\and
\IEEEauthorblockN{Jawad Chowdhury}
\IEEEauthorblockA{\textit{Department of Computer Science} \\
\textit{UNC Charlotte}\\
Charlotte, USA \\
mchowdh5@uncc.edu}
\and
\IEEEauthorblockN{Gabriel Terejanu}
\IEEEauthorblockA{\textit{Department of Computer Science} \\
\textit{UNC Charlotte}\\
Charlotte, USA \\
gabriel.terejanu@uncc.edu}
}

\maketitle

\begin{abstract}
Causal structure learning from observational data remains a non-trivial task due to various factors such as finite sampling, unobserved confounding factors, and measurement errors. Constraint-based and score-based methods tend to suffer from high computational complexity due to the combinatorial nature of estimating the directed acyclic graph (DAG). Motivated by the `Cause-Effect Pair' NIPS 2013 Workshop on Causality Challenge, in this paper, we take a different approach and generate a probability distribution over all possible graphs informed by the cause-effect pair features proposed in response to the workshop challenge. The goal of the paper is to propose new methods based on this probabilistic information and compare their performance with traditional and state-of-the-art approaches. Our experiments, on both synthetic and real datasets, show that our proposed methods not only have statistically similar or better performances than some traditional approaches but also are computationally faster.
\end{abstract}


%-------------INTRODUCTION-----------------------------------------
\section{Introduction}
\label{sec-introduction}

Machine learning methods, deep learning in particular, have achieved unparalleled predictive performance in the past two decades. Nevertheless, these correlation-based models exhibit significant limitations when applied to out-of-distribution data and prescriptive analytics, which is grounded in causal inference. Learning the underlying causal structure is an important task to both constrain a model by reducing  spurious correlations~\cite{lake2017building} and perform What-If analysis~\cite{pearl-causality}. Learning causal relationships from observational data, also known as causal discovery, remains an active and challenging research topic~\cite{pearl-causality,pc-paper,peters2017elements}.

Several causal discovery methods have been proposed in the literature. Constraint-based approaches learn the causal skeleton using conditional independence test using the joint probability distribution of the data and identify edge directions up to their Markov equivalence class~\cite{pc-paper,pearl1995theory,colombo2012order,sun2007kernel,zhang2012kernel}. 

Score-based approaches learn the causal graph $\mathcal{G}$ by optimizing a score function generally computed with respect to observational data~\cite{ges-paper,ramsey2017million,nandy2018high}. Unfortunately, these methods suffer from super-exponential computational complexity in the number of nodes. Tsamardinos et al. \cite{tsamardinos2006max} propose a hybrid method where they use a constraint-based approach to reduce the search space in score-based methods. However, this method relies on local heuristics and lacks a standard way of choosing score functions and search strategies~\cite{zhu2019causal}.

A promising direction, NOTEARS \cite{notears2018}, formulates a smooth characterization of acyclicity that can be incorporated into a continuous optimization and solved using well-known numerical methods. NOTEARS was later extended to parametric nonlinear models and nonparametric models \cite{notears}. GOLEM\cite{golem} also adopts a continuous optimization framework, however, it makes use of a linear DAG learning model and doesn't capture non-linear relationships.
% {\color{red}Another work by  GOLEM\cite{golem} adopts a continuous optimization framework in the linear case for learning DAG models where they have studied the usage of empirical correlation and covariance matrices. However, it is a linear DAG learning model and doesn't provide information with non-linear relationships.} 
A different approach looks at identifying cause-effect pairs using the statistical techniques from observational data~\cite{guyon2019evaluation,fonollosa2019conditional}. Singh et al.\cite{singh2017deep} use deep convolutional neural network (CNN) models to determine the directions of pairwise causal edges from observational data. Hassanzadeh et al.\cite{hassanzadeh2019answering} formulate the pairwise causal discovery techniques as binary causal problems where they try to answer if there exist any causal relations between two variables in the context of Natural Language Processing (NLP). Nevertheless, they have not studied how the predicted edge directions can be used to provide a solution to causal graph identification.

Motivated by applications in biological networks, Medvedovsky
et al.\cite{medvedovsky2008algorithm} propose an approximation algorithm to orient a graph by maximizing the number of pairs that admit a directed path from known pairs of sources and targets. Nevertheless, given the peculiarities of the application based on prior knowledge constrains, this approach falls short from identifying the entire causal graph. Therefore, discovering the causal graph, including pairwise causal relations from observational data remains a challenging task due to various factors such as finite sampling and measurement errors. In general, identifying cause-effect relationships requires controlled experimentation which is expensive and/or technically and ethically impossible to perform~\cite{stegle2010probabilistic}.

We propose a probabilistic approach to discover causal structures using the cause-effect pairs features proposed in response to the `Cause-Effect Pair' at the NIPS 2013 Workshop on Causality challenge. The paper introduces the following novel contributions: (1) generate a probability distribution over all the edges of a digraph using various statistical and information-theoretic features that describe the relationships between any two variables in the dataset; (2) generate the most likely probability distribution of directed acyclic graphs (DAG) using the maximum spanning DAG; (3) generate an approximate solution to the causal graph problem by estimating the digraph and DAG using maximum likelihood estimate with the probability distributions in (1) and (2) respectively; (4) overall, our proposed methods are comparable with traditional ones (PC, GES), while benefiting from polynomial time complexity as compared to super-exponential time complexity; and (5) finally, by comparing with state-of-the-art methods such as NOTEARS-MLP, we show that future improvements are possible by further leveraging global graph information.

Section~\ref{sec-methodology} introduces the problem formulation and details our causal discovery approach based on causal-effect pairs. The empirical evaluation of our methods is described in Section~\ref{sec-experiments}. Lastly, in Section~\ref{conclusions} we present our findings in brief and opportunities for future improvements.

% METHODOLOGY % --------------------------------------------------------------
\section{Methodology}
\label{sec-methodology}
% \label{sec-problem form}

Given $n$ i.i.d. observations in the data matrix $\mathbf{X} = [\mathbf{x_1}\ldots\mathbf{x_d}] \in \mathbb{R}^{n\times d}$, the goal of causal discovery is to estimate the underlying causal relations encoded by the directed acyclic graph (DAG), $\mathcal{G}_{\text{DAG}}=(V,E)$. $V$ comprises of nodes corresponding to the observed random variables $X_i$ for $i=1 \ldots d$ and the edges in $E$ correspond to the causal relations encoded by $\mathcal{G}_{\text{DAG}}$. Namely, the existence of the edge $i \to j$ corresponds to a direct causal relationship between $X_i$ (the cause) and $X_j$ (the effect).

Our approach is to leverage the work on cause-effect pairs which uses machine learning to predict the probability distribution $p(e_{ij}|f)$ of causal relation between two variables $X_i$ and $X_j$ given the observational dataset $[\mathbf{x_i}, \mathbf{x_j}] \in \mathbb{R}^{n\times 2}$. 
\begin{eqnarray}
    \label{eq:model}
    p(e_{ij}|f) &=& f([\mathbf{x_i}, \mathbf{x_j}]),~\text{for}~i<j
\end{eqnarray}

Here, we assume that $f(\cdot)$ is a pre-trained machine learning model and $e_{ij} \in [-1, 0, 1]$.

\begin{equation}
\nonumber
\label{eq-causal label}
e_{ij} = 
	\begin{cases}
	-1: &\text{$j \to i$, causal relation exists from $X_j$ to $X_i$}\\
	\;\;\:0: &\text{$i\not\to j$ and $j\not\to i$, no direct causal relation}\\
	&\text{between $X_i$ and $X_j$}\\
	\;\;\:1: &\text{$i \to j$ causal relation exists from $X_i$ to $X_j$}
	\end{cases}
\end{equation}

After calculating the probability distributions of causal relations between all the pairs in the dataset, a naive approach to construct the probability distribution of a digraph $\mathcal{G}$ is to assume that the causal-pairs are independent. In Section \ref{sec-method2.2}, we show that our proposed approach of enforcing DAGness does correlate these causal-pairs and provides us with additional global information to constrain the graph
probability distribution and allows us to move beyond the initial edge independence.
%
% {\color{red}It is to mention that we leveraged the independence of the causal-pairs and used a naive approach to build the probability distribution over the causal graphs which helps in polynomial time computation. But later in Section \ref{sec-method2.2}, we show enforcing DAGness correlates these causal-pairs and provides us
% with additional global information to constrain the graph
% probability distribution and allows us to move beyond the
% initial edge independence.} 
%
\begin{equation}
\label{eq:PG}
    p(\mathcal{G}|f) = \prod_{i<j} p(e_{ij}|f) 
\end{equation}
Given this rich probabilistic information on all the causal relationships in the dataset, one may choose to generate the maximum likelihood digraph.
\begin{equation}
\label{eq:MLG}
    \mathcal{G}_\text{ML} = \arg\max_\mathcal{G} p(\mathcal{G}|f)  
\end{equation}
Note that the samples from the probability distribution, Eq.~\ref{eq:PG}, and the maximum likelihood estimate, Eq.~\ref{eq:MLG}, are digraphs with no guarantees that they are acyclic. In the following, we propose to generate the most likely probability distribution of directed acyclic graphs (DAG) using the maximum spanning DAG approach~\cite{schluter-2014-maximum} as well as estimate a representative DAG using the maximum likelihood estimate.

\subsection{Developing causal-pair models}
\label{sec-method2.1}

The model $f(\cdot)$ in Eq.~\ref{eq:model} can be trained using synthetic datasets or real datasets with known causal relations. Given a set of labeled datasets $\{ ([\mathbf{x_i}, \mathbf{x_j}], y_{ij})_k \}$, in this paper we take the approach of engineering features from this dataset using various statistical and information-theoretic measures such as: minimum or maximum value of a variable; number of unique samples of a variable; entropy, mutual information, uniform divergence; slope-based information geometric causal inference (IGCI), Hilbert Schmidt independence criterion (HSIC); Pearson R coefficient; Spearman's rank coefficient; moments and mixed moments such as skewness and kurtosis. Therefore, the machine learning model $f(\cdot)$ can be trained on the new engineered dataset with  features previously introduced. We note that our proposed methodology is agnostic to the features deployed and it works with any causal-pairs model. Also, the computational runtime of calculating any of these features such as HSIC, Pearson R coefficient has no effect asymptotically on the computational complexity - it just increases the constant of the polynomial runtime. We also note that additional improvements might be brought by developing deep neural network architectures capable to extract informative representations for predicting the target $y_{ij} \in [-1, 0, 1]$ directly from the sample dataset, but it is left as future work.

\subsection{Enforcing DAGness}
\label{sec-method2.2}
In this section, we propose to derive a probability distribution that guarantees that the sample graphs are DAGs. This takes the form of the  probability distribution in Eq.~\ref{eq:PDAG} which unlike Eq.~\ref{eq:PG} contains the DAGness condition.
\begin{equation}
\label{eq:PDAG}
    p(\mathcal{G}|f, \text{DAG}) = \sum_\pi p(\mathcal{G}|f, \text{DAG}, \pi) p(\pi|f)
\end{equation}
Due to computational intractability, we have chosen to build this conditional distribution not using the Bayes rule and utilizing Eq.~\ref{eq:PG} as prior, but rather as the law of total probability where we integrate out the topological ordering $\pi$ of the vertices. We note that for a topological ordering where node $i$ comes before node $j$, it implies that a directed edge can only happen from $i$ to $j$. We also note that the possibility of no causal relation between $i$ and $j$ is not excluded in this context. Both causal, $i\rightarrow j$, and noncausal, $i\not\to j$ and $j\not\to i$, are possible.

To generate a representative DAG one can use the maximum likelihood estimate. This however is intractable, and we do assume that there is a topological sorting of vertices that also covers our maximum likelihood DAG.
\begin{eqnarray}
    \mathcal{G}_\text{DAG} &=& \arg\max_\mathcal{G} p(\mathcal{G}|f, \text{DAG}) \\
    &\approx&  \arg\max_\mathcal{G} p(\mathcal{G}|f, \text{DAG}, \pi_\text{ML}) \label{eq:MLDAG}
\end{eqnarray}
We propose to approximate the topological ordering, $\pi_\text{ML}$, by the topological sorting of the Maximum Spanning DAG (MSDAG)~\cite{schluter-2014-maximum} of the induced weighted graph by the probability of causal relations. 
\begin{eqnarray}
    \pi_\text{ML} &=& \arg\max_\pi p(\pi|f) \\
    &\approx& \text{toposort(MSDAG($\mathcal{G}_A$))}
\end{eqnarray}
We build the following weighted adjacency matrix $A \in \mathcal{R}^{d \times d}$, which contains the probability of all directed edges as weights.
\begin{equation}
\label{eq-matrixconv}
\begin{aligned}
    &A[i,j] = p(e_{ij}=1|f) \\
    &A[j,i] = p(e_{ij}=-1|f)  
\end{aligned}
\end{equation}
%
%Note that the weights in the matrix $A$ represent the probabilities of causal relations and the probability of no causal relations is implicitly defined as $p(e_{ij}=0) = 1 - A[i,j] - A[j,i]$.
%
Let $\mathcal{G}_A$ be a weighted graph induced by the adjacency matrix $A$. The goal is to find the topological sorting of the MSDAG of $\mathcal{G}_A$. The motivation is to accommodate as many directed edges with large probabilities as possible. We use the approach introduced by \cite{mcdonald-pereira-2006-online} to approximate the MSDAG by first constructing the maximum spanning tree and greedily adding edges in the descending order of the weights as long as no cycles are formed. Note that a topological sorting derived this way still accommodates the possibility of no edges to account for their probability in the maximum likelihood DAG, Eq.~\ref{eq:MLDAG}.
\begin{equation}
\label{eq:PDAG_new}
    p(\mathcal{G}|f, \text{DAG}, \pi_\text{ML}) = \prod_{\pi^{-1}_\text{ML}[i] < \pi^{-1}_\text{ML}[j]} p(e_{i\rightarrow j}|f) 
\end{equation}
Given the maximum likelihood of topological ordering, one can easily calculate the probability distribution in Eq.~\ref{eq:MLDAG} by constraining the direction of the edge based on the node ordering, see Eq.~\ref{eq:PDAG_new}. In this context, we are left with only two possibilities when node $i$ appears before $j$ in $\pi_\text{ML}$. Either there is an edge from $i$ to $j$ or there is no edge between them.
\begin{equation}
\nonumber
\label{eq-causal label}
e_{i\rightarrow j} = 
	\begin{cases}
	1: &\text{$i \to j$, causal relation exists from $X_i$ to $X_j$}\\
	0: &\text{$i\not\to j$ and $j\not\to i$, no direct causal relation}\\
	&\text{between $X_i$ and $X_j$}
	\end{cases}
\end{equation}
By constraining the direction of edges we need to re-normalizing the edge  probabilities as follows.
\begin{equation}
    p(e_{i\rightarrow j}=1|f) = \frac{p(e_{ij}=1|f)}{p(e_{ij}=1|f)+p(e_{ij}=0|f)}
\end{equation}
Enforcing DAGness in Eq.~\ref{eq:PDAG_new} and consequently Eq.~\ref{eq:MLDAG} provides us with additional global information to constrain the graph probability distribution and allows us to move beyond the initial edge independence in Eq.~\ref{eq:PG} and Eq.~\ref{eq:MLG} respectively, which was derived from just pair-wise (local) information.

% EXPERIMENT % -------------------------------------------------------------------
\section{Experiments}
\label{sec-experiments}

We have used the following labels for our approaches: PG given by Eq.~\ref{eq:PG}, MLG given by Eq.~\ref{eq:MLG}, PDAG given by Eq.~\ref{eq:PDAG_new}, and MLDAG given by Eq.~\ref{eq:MLDAG}. We show the empirical results of our approaches applied on both synthetic and real-world datasets and we compare our performance with two traditional approaches: the PC algorithm~\cite{pc-paper} and the GES algorithm~\cite{ges-paper} and with a state-of-the-art approach, NOTEARS-MLP~\cite{notears}. Both PC and GES have public implementations and we used the code\footnote{PC: https://github.com/keiichishima/pcalg\label{fn-pc}}\textsuperscript{,}\footnote{GES: https://github.com/juangamella/ges\label{fn-ges}} from their git repositories. In addition, for the NOTEARS-MLP\cite{notears} approach, we followed the implementation stated in the paper and git repository\footnote{NOTEARS-MLP: https://github.com/xunzheng/notears\label{fn-notears}}. We used the default settings and default hyper-parameters for all these three implementations. However, Reisach et. al. \cite{beware-dag} in their paper highlight that continuous score-based approaches i.e. NOTEARS-MLP\cite{notears} in particular suffer highly from data scaling which was addressed from a theoretical perspective. Therefore, we standardize the features for both the synthetic and real-world datasets by removing the mean and scaling to unit variance.

% DATA SET % ------------------------------------------------------
\subsection{Data Sets}

\paragraph{Cause-Effect Pair Train Data} To train our model, we have used the Cause-effect pairs dataset\footnote{Cause-Effect Pair Dataset: https://www.kaggle.com/c/cause-effect-pairs/data} from the NIPS 2013 Workshop on Causality. The train data contains 4050 samples of pairs (attribute A and attribute B) of real variables with known causal relationships and every node within a pair has the same number of data samples. In addition, these known ground truths are derived from expert domains such as chemistry, ecology, engineering, medicine, physics, sociology, etc., and these pairs are intermixed with controls such as pairs of independent variables and pairs of dependent variables but not causally related. 

\paragraph{Synthetic Test Data} To evaluate the performance of our methods on causal graph estimation, we have generated synthetic data for testing. We have considered 16 types of different graph combinations having similar criteria: number of nodes, $d=[10,20]$, number of edges, $e=[1d, 4d]$, number of data samples per node, $n=[200,1000]$, and graph models from Erdos-Renyi(ER) and Scale-Free (SF). We have generated non-linear data samples for the graph nodes similar to data generation utilities available in the NOTEARS-MLP implementation. In addition, for each of these 16 graph types, we have generated 10 random graph structures with ground truths to test our methods. The outputs are then summarized over these 10 graph structures to report our results for all 16 graph combinations.

% METRICS % --------------------------------------------------------------------------

\subsection{Metrics}We consider three performance metrics to evaluate the causal graphs: True Positive Rate (TPR), False Positive Rate (FPR) and Structural Hamming Distance (SHD). A lower SHD and FPR indicate a better performance whereas a higher TPR is better. However, since both PC and GES may generate outputs with undirected edges, we treated an undirected edge as a true edge with a probability of $0.5$ and a false edge with the same probability. SHD, TPR, and FPR were implemented by their definition for PC, GES, NOTEARS-MLP, and our two maximum likelihood estimates (MLG given by Eq.~\ref{eq:MLG} and MLDAG given by Eq.~\ref{eq:MLDAG}).

As for our probabilistic approaches (PG given by Eq.~\ref{eq:PG} and PDAG given by Eq.~\ref{eq:PDAG_new}), given a true graph $\mathcal{G}_{true}$ and an adjacency matrix $\mathcal{A}$ with edge probabilities, we calculate SHD using Eq.~\ref{eq-shd3p}, TPR using Eq.~\ref{eq-tpr3p} and FPR using Eq.~\ref{eq-fpr3p}.  
\begin{equation}
\small
\begin{aligned}
SHD=&\sum_{\substack{{i < j}\\{(i,j)\in E(\mathcal{G}_{true})}\\{(j,i)\not\in E(\mathcal{G}_{true})}}}(1-\mathcal{A}[i,j])
+\sum_{\substack{{i < j}\\{(i,j)\not\in E(\mathcal{G}_{true})}\\{(j,i)\in E(\mathcal{G}_{true})}}}(1-\mathcal{A}[j,i])\\
&+\sum_{\substack{{i < j}\\{(i,j)\not\in E(\mathcal{G}_{true})}\\{(j,i)\not\in E(\mathcal{G}_{true})}}}(\mathcal{A}[i,j]+\mathcal{A}[j,i])
\label{eq-shd3p}
\end{aligned}
\end{equation}
\begin{equation}
\small
\begin{aligned}
TP=\sum_{\substack{{i < j}\\{(i,j)\in E(\mathcal{G}_{true})}\\{(j,i)\not\in E(\mathcal{G}_{true})}}}\mathcal{A}[i,j]+\sum_{\substack{{i < j}\\{(i,j)\not\in E(\mathcal{G}_{true})}\\{(j,i)\in E(\mathcal{G}_{true})}}}\mathcal{A}[j,i]
\end{aligned}
\end{equation}
\begin{equation}
\small
\label{eq-tpr3p}
TPR = \frac{TP}{max(|E(\mathcal{G}_{true})|, 1)}
\end{equation}

\begin{equation}
\small
\begin{aligned}
FP=&\sum_{\substack{{i < j}\\{(i,j)\in E(\mathcal{G}_{true})}\\{(j,i)\not\in E(\mathcal{G}_{true})}}}\mathcal{A}[j,i]+\sum_{\substack{{i < j}\\{(i,j)\not\in E(\mathcal{G}_{true})}\\{(j,i)\in E(\mathcal{G}_{true})}}}\mathcal{A}[i,j]\\
&+\sum_{\substack{{i < j}\\{(i,j)\not\in E(\mathcal{G}_{true})}\\{(j,i)\not\in E(\mathcal{G}_{true})}}}(\mathcal{A}[i,j]+\mathcal{A}[j,i])
\end{aligned}
\end{equation}
\begin{equation}
\small
\label{eq-fpr3p}
FPR = \frac{FP}{max((M-|E(\mathcal{G}_{true})|), 1)}
\end{equation}

\noindent
Here, $M=\frac{d(d-1)}{2}$ is the number of possible edges of the graph $\mathcal{G}_{true}$ and $d$ is the total number of nodes in the graph.

Note that since we are calculating these metrics over 160 different graph structures of various sizes in our test data, we report a normalized SHD over the number of graph nodes (SHD/d) as our SHD measure. TPR and FPR are normalized by definition.

% MLG (Eq.~\ref{eq:MLG}) 
% MLDAG (Eq.~\ref{eq:MLDAG})

% PG (Eq.~\ref{eq:PG}) 
% PDAG (Eq.~\ref{eq:PDAG_new})

%table 1 pair data%---------ER GRAPH----------------------------------

% \usepackage{booktabs}
\begin{table*}[t]
\caption{Edge probability model trained on cause-effect pairs data provided at the NIPS 2013 Workshop on Causality. The means and standard errors of the performance metrics are based on the 80 Erdos-Renyi (ER) graph structures in the test data.}
\label{table-pair data-ER}
\vskip 0.15in
\begin{center}
\begin{small}
\begin{sc}
\begin{tabular}{@{}|l|r|r|r|r|r|r|r|@{}}
\hline
% & A & B & C & D & E & F & G \\ \midrule
Metrics & \begin{tabular}[c]{@{}r@{}}PG (Eq.~\ref{eq:PG})\end{tabular} & \begin{tabular}[c]{@{}r@{}}MLG (Eq.~\ref{eq:MLG})\end{tabular} & \begin{tabular}[c]{@{}r@{}}PDAG (Eq.~\ref{eq:PDAG_new})\end{tabular} & \begin{tabular}[c]{@{}r@{}}MLDAG (Eq.~\ref{eq:MLDAG})\end{tabular} & PC & GES & NOTEARS-MLP \\ \hline
SHD/$d$ & 2.38$\pm$0.14 & 2.32$\pm$0.17 & 2.30$\pm$0.15 & 2.18$\pm$0.16 & 2.40$\pm$0.21 & 1.78$\pm$0.13 & 1.33$\pm$0.10 \\
TPR & 0.39$\pm$0.02 & 0.15$\pm$0.02 & 0.38$\pm$0.02 & 0.28$\pm$0.02 & 0.17$\pm$0.02 & 0.48$\pm$0.02 & 0.58$\pm$0.02 \\
FPR & 0.72$\pm$0.10 & 0.07$\pm$0.01 & 0.61$\pm$0.09 & 0.29$\pm$0.05 & 0.22$\pm$0.04 & 0.87$\pm$0.15 & 0.32$\pm$0.06 \\ \hline
\end{tabular}
\end{sc}
\end{small}
\end{center}
\vskip -0.1in
\end{table*}

%table 2 pair data%---------SF GRAPH----------------------------------

% \usepackage{booktabs}
\begin{table*}[t]
\caption{Edge probability model trained on cause-effect pairs data provided at the NIPS 2013 Workshop on Causality. The means and standard errors of the performance metrics are based on the 80 Scale-Free (SF) graph structures in the test data.}
\label{table-pair data-SF}
\vskip 0.15in
\begin{center}
\begin{small}
\begin{sc}
\begin{tabular}{@{}|l|r|r|r|r|r|r|r|@{}}
\hline
% & A & B & C & D & E & F & G \\ \midrule
Metrics & \begin{tabular}[c]{@{}r@{}}PG (Eq.~\ref{eq:PG})\end{tabular} & \begin{tabular}[c]{@{}r@{}}MLG (Eq.~\ref{eq:MLG})\end{tabular} & \begin{tabular}[c]{@{}r@{}}PDAG (Eq.~\ref{eq:PDAG_new})\end{tabular} & \begin{tabular}[c]{@{}r@{}}MLDAG (Eq.~\ref{eq:MLDAG})\end{tabular} & PC & GES & NOTEARS-MLP \\ \hline
SHD/$d$ & 2.02$\pm$0.12 & 1.97$\pm$0.13 & 1.96$\pm$0.12 & 1.88$\pm$0.13 & 1.93$\pm$0.15 & 1.43$\pm$0.11 & 1.36$\pm$0.11 \\
TPR & 0.31$\pm$0.01 & 0.12$\pm$0.01 & 0.30$\pm$0.01 & 0.20$\pm$0.01 & 0.17$\pm$0.02 & 0.51$\pm$0.03 & 0.47$\pm$0.02 \\ 
FPR & 0.26$\pm$0.02 & 0.03$\pm$0.01 & 0.21$\pm$0.02 & 0.09$\pm$0.01 & 0.08$\pm$0.01 & 0.26$\pm$0.04 & 0.12$\pm$0.02 \\ \hline
\end{tabular}
\end{sc}
\end{small}
\end{center}
\vskip -0.1in
\end{table*}

% SIMULATION % -------------------------------------------------------------------------
\subsection{Simulation} For our implementation, we first extract features from the data-pairs using the feature extraction method of Team-Jarfo\cite{jarfo2016}, the second winner from the NIPS 2013 Workshop on Causality challenge. We have also used the causal pairs
model of the third winning team from the above-mentioned competition and we have found that
it doesn't have better performance than Team-Jarfo\cite{jarfo2016} as expected. We couldn’t run a performance analysis with the model proposed by the No.1 team due to the lack of availability of their implemented code. We have extracted 130 features from the pairs for all training and testing datasets using the code\footnote{Team-Jarfo Code: https://github.com/jarfo/cause-effect\label{fn-jarfo}} of Team-Jarfo\cite{jarfo2016}. This feature set contains some standard statistical features as well as new measures based on variable measures of the conditional distribution. We train a multi-classifier model based on LightGBM~\cite{lightgbm-microsoft} which is a Gradient Boosting Decision Tree (GBDT) algorithm developed by Microsoft. LightGBM speeds up the training process by using a histogram-based algorithm~\cite{lgbm-histo1,lgbm-histo2} and combines weak learners into strong ones using an iterative approach~\cite{lgbm-iterative} to optimize parallel learning. We create the LightGBM classifier from the Python library PyCaret~\cite{PyCaret} and select the hyper-parameters by tuning the model using the PyCaret `tune\_model()' function optimized over AUC. 

We train our classifier model on cause-effect pairs data using the 130 extracted features and make predictions on the synthetic testing data. We predict a probability distribution over the three classes of edge directions (backward edge, no edge, forward edge) for all edges in the testing dataset. 
Finally, using the methods described in Section~\ref{sec-methodology} we calculate the metrics of our causal graph estimation methods from the predicted probability distribution and compare the results with the benchmark approaches.

%Observation------------------------------------

Table~\ref{table-pair data-ER} and Table~\ref{table-pair data-SF} show the empirical results of our methods applied to 80 different Erdos-Renyi graph structures and 80 different Scale-Free graphs, respectively. From these two tables, we see that estimating causal graphs is more challenging for Erdos-Renyi graph structures than Scale-Free graphs for all the methods. We also note that the performance of NOTEARS-MLP is superior to all other methods in terms of SHD and TPR in particular and it does have implications for further developing our proposed methods as detailed later.

Regardless of the graph structure, we make the following observations. (1) PG performs better than PC in terms of SHD and TPR and is better than GES in terms of FPR. (2) MLG does improve the FPR but at the cost of degrading the TPR. (3) We do see a significant improvement by enforcing DAGness. Namely, PDAG does improve FPR over PG, but not sufficient to be statistically better/similar to PC. (4) However, this does happen when we take the maximum likelihood of the conditional probability. Namely, MLDAG performs at the same FPR level as PC.

From these results, it becomes clear that using global information by constraining the graphs to be DAGs does improve the performance compared to using the pair-wise probabilities naively. However, when we compare it with NOTEARS-MLP, it also becomes clear that this is not sufficient and that the next iteration of methods needs to develop features that intrinsically exploit global information.

Nevertheless, one of the distinguishable advantages our methods have over PC and GES is that they not only perform statistically better/similar but also they have significantly low computational complexity. While both PC and GES have exponential time complexity due to their combinatorial approach, our methods run in polynomial time $\mathcal{O}(d^2)$ in the number of nodes $d$ as they exploit local node pairs information.

% REAL WORLD DATA % ---------------------------------------------------------

\subsection{Real-World Data}
For real data, we consider the dataset published by \cite{sachs-paper}, which is based on the expression level of proteins. This protein network is largely used  in the scientific community as a real application due to the consensus ground truths of the graph structure. It has $11$ different protein cells which are considered as graph nodes $d$. We considered $e=17$ edges as ground truths representing the protein signaling network. We aggregated the $9$ different data files, resulting in a sample size $n=7466$ in our experiment.

%% protein data table CE pairs-----------------------------------

% MLG (Eq.~\ref{eq:MLG}) 
% MLDAG (Eq.~\ref{eq:MLDAG})

% PG (Eq.~\ref{eq:PG}) 
% PDAG (Eq.~\ref{eq:PDAG_new})

% \usepackage{booktabs}
\begin{table*}[htbp]
\caption{Comparison of our probabilistic methods with GES and NOTEARS-MLP that were applied on protein network dataset using cause-effect pairs as training data.}
\label{table-protein data}
\vskip 0.15in
\begin{center}
\begin{small}
\begin{sc}
\begin{tabular}{@{}|l|r|r|r|r|r|r|@{}}
\hline
Metrics & \begin{tabular}[c]{@{}r@{}}PG (Eq.~\ref{eq:PG})\end{tabular} & \begin{tabular}[c]{@{}r@{}}MLG (Eq.~\ref{eq:MLG})\end{tabular} & \begin{tabular}[c]{@{}r@{}}PDAG (Eq.~\ref{eq:PDAG_new})\end{tabular} & \begin{tabular}[c]{@{}r@{}}MLDAG (Eq.~\ref{eq:MLDAG})\end{tabular} & GES & NOTEARS-MLP \\ \hline
Predicted Edges & 36.14 & 9.82 & 33.16 & 18.48 & 34 & 42.23 \\ 
Correct Edges & 6.7 & 3.04 & 7.42 & 4.91 & 5.5 & 5.83 \\ 
Reversed Edges & 7.77 & 4.26 & 6.62 & 5.41 & 9.5 & 7.18 \\
\hline
\end{tabular}
\end{sc}
\end{small}
\end{center}
\vskip -0.1in
\end{table*}

%% protein data table NOTEARS ORIGINAL Results-----------------------------------

% \usepackage{booktabs}
\begin{table*}[htbp]
\caption{Comparison of our probabilistic methods with GES and NOTEARS-MLP (results reported from the original manuscript\cite{notears}). These methods were applied on non-standardized protein network dataset.}
\label{table-four}
\vskip 0.15in
\begin{center}
\begin{small}
\begin{sc}
\begin{tabular}{@{}|l|r|r|r|r|r|r|@{}}
\hline
Metrics & \begin{tabular}[c]{@{}r@{}}PG (Eq.~\ref{eq:PG})\end{tabular} & \begin{tabular}[c]{@{}r@{}}MLG (Eq.~\ref{eq:MLG})\end{tabular} & \begin{tabular}[c]{@{}r@{}}PDAG (Eq.~\ref{eq:PDAG_new})\end{tabular} & \begin{tabular}[c]{@{}r@{}}MLDAG (Eq.~\ref{eq:MLDAG})\end{tabular} & GES & NOTEARS-MLP \\ \hline
Predicted Edges & 38.01 & 10.41 & 34.81 & 20.60 & 34 & 13 \\ 
Correct Edges & 6.21 & 1.52 & 6.47 & 4.71 & 5.5 & 7 \\ 
Reversed Edges & 8.26 & 4.04 & 7.49 & 6.32 & 9.5 & 3 \\
\hline
\end{tabular}
\end{sc}
\end{small}
\end{center}
\vskip -0.1in
\end{table*}

Unlike synthetic data sets, we have considered three different metrics for this protein dataset: the total number of predicted edges, the number of correct edge predictions, and the number of reversed edge predictions. Since this protein network uses a consensus over the number of true edges ($17$ known edges as ground truth), we do not know the actual true graph for the entire network. Therefore, a metric such as SHD becomes meaningless. Furthermore, since GES algorithms generate graphs that contain bidirectional edges, similarly to the synthetic results, we considered the bidirectional edge as increasing the number of corrected edges with $0.5$ and the number of reversed edges with $0.5$ as well.

In Table~\ref{table-protein data}, we show the performance evaluation of our methods applied to the protein network dataset. We note that both PG and PDAG perform better than GES and NOTEARS-MLP in terms of predicting the number of correct edges. In addition, we find that taking the maximum likelihood estimates, MLG and MLDAG, significantly reduce the number of false edge predictions but their performances in predicting correct edges also degrade. The degradation is less severe between PDAG and MLDAG than PG and MLG.

PG, PDAG, as well as GES have the best performance in terms of the total number of predicted edges than NOTEARS-MLP, but MLDAG has the best total number of predicted edges which is close to the number of true edges in the dataset. We also note that MLG and MLDAG have better performance than NOTEARS-MLP in terms of fewer reversed edges. Finally, PGDAG shows an overall better performance than PG, further demonstrating the impact of enforcing DAGness. 

\textbf{Sensitivity to data scaling}. Table~\ref{table-four} shows the results of our methods as well as GES and NOTEARS-MLP~\cite{notears} applied on protein dataset before scaling them. We observe that our methods: PG, PDAG, MLG, MLDAG have almost similar results in both Table~\ref{table-protein data} (after data scaling) and Table~\ref{table-four} (before data scaling). GES is not affected by the data scaling. However, as expected, we see that NOTEARS-MLP has very different results in these two tables where it suffers highly from data-scaling, which is consistent with the sensitivity to scaling results in Ref.\cite{beware-dag}. It is to mention that in Table~\ref{table-four}, NOTEARS-MLP results are reported from the original manuscript~\cite{notears},  whereas in Table~\ref{table-protein data}, we use the same implementation\textsuperscript{\ref{fn-notears}} of NOTEARS-MLP~\cite{notears} with default parameters and applied on the standardized protein dataset by removing the mean and scaling to unit variance.

% DISCUSSION % --------------------------------------------------------
\section{Conclusions}
\label{conclusions}
In this paper, we have introduced a novel approach to causal discovery by leveraging the probabilistic information of pairwise causal edges. We have proposed to go beyond the naive approach to generate graph probabilities from causal pair probabilities by enforcing the graph to be acyclic and approximating its solution using the maximum spanning directed acyclic graph approach. Enforcing acyclicity clearly improves the performance on both synthetic and real datasets compared with the naive approach.

We have shown that our methods have statistically better and/or similar performances than some traditional methods. More importantly, this performance comes with just polynomial run-time as compared with the exponential run-time of traditional methods that are combinatorial in nature. These promising results prompt us to further look into improving the causal pair feature generation to intrinsically capture global information.

% \bibliographystyle{IEEEtran}
% \bibliography{references}

% DISCUSSION % --------------------------------------------------------
\section*{Acknowledgement}
Research was sponsored by the Army Research Office and was
accomplished under Grant Number W911NF-22-1-0035. The views and conclusions contained in this document are those of the authors and should not be interpreted as representing the official policies, either
expressed or implied, of the Army Research Office or the U.S. Government. The U.S. Government is
authorized to reproduce and distribute reprints for Government purposes notwithstanding any copyright
notation herein.

\end{document}